\documentclass[acmsmall]{acmart}

\usepackage{amsmath,amsfonts}
\usepackage{algorithmic}
\usepackage{graphicx}
\usepackage{booktabs}
\usepackage{multirow}
\usepackage{textcomp}
\usepackage{xcolor}
\usepackage{booktabs}
\usepackage{url}
\usepackage[ruled, linesnumbered]{algorithm2e}
\usepackage{caption}
\usepackage{subcaption}
\usepackage[inline]{enumitem}

\AtBeginDocument{%
  \providecommand\BibTeX{{%
    \normalfont B\kern-0.5em{\scshape i\kern-0.25em b}\kern-0.8em\TeX}}}

\setcopyright{acmcopyright}
\makeatletter
\newcommand{\inlineitem}[1][]{%
\ifnum\enit@type=\tw@
    {\descriptionlabel{#1}}
  \hspace{\labelsep}%
\else
  \ifnum\enit@type=\z@
       \refstepcounter{\@listctr}\fi
    \quad\@itemlabel\hspace{\labelsep}%
\fi}
\makeatother

\newcommand{\skm}[1]{\textcolor{black}{#1}}
\newcommand{\YC}[1]{\textcolor{black}{#1}}

\newcommand{\BT}[1]{\textcolor{black}{#1}}
\newcommand{\NC}[1]{\textcolor{black}{#1}}
\newcommand{\rev}[1]{\textcolor{black}{#1}}
\newcommand{\camready}[1]{\textcolor{black}{#1}}

\begin{document}


\title{SIAM: Chiplet-based Scalable In-Memory Acceleration \\ with Mesh for Deep Neural Networks}

\thanks{This article appears as part of the ESWEEK-TECS special issue and was presented in the International Conference on Hardware/Software Codesign and System Synthesis (CODES+ISSS), 2021.}

\author{Gokul Krishnan}
\email{gkrish19@asu.edu}
\orcid{0000-0003-1085-2189}
\affiliation{%
	\institution{Arizona State University}
	\department{School of Electrical, Computer, and Energy Engineering}
	\city{Tempe}
	\state{AZ}
	\postcode{85287}
	\country{USA}
}

\author{Sumit K. Mandal}
\email{skmandal@wisc.edu}
\orcid{0000-0003-1085-2189}
\affiliation{%
	\institution{University of Wisconsin-Madison}
	\department{Department of Electrical and Computer Engineering}
	\city{Madison}
	\state{WI}
	\postcode{53706}
	\country{USA}
}

\author{Manvitha Pannala}
\email{mpannal1@asu.edu}
\orcid{}
\affiliation{%
	\institution{Arizona State University}
	\department{School of Electrical, Computer, and Energy Engineering}
	\city{Tempe}
	\state{AZ}
	\postcode{85287}
	\country{USA}
}

\author{Chaitali Chakrabarti}
\email{chaitali@asu.edu}
\orcid{0000-0003-1085-2189}
\affiliation{%
	\institution{Arizona State University}
	\department{School of Electrical, Computer, and Energy Engineering}
	\city{Tempe}
	\state{AZ}
	\postcode{85287}
	\country{USA}
}

\author{Jae-sun Seo}
\email{jseo28@asu.edu}
\orcid{0000-0003-1085-2189}
\affiliation{%
	\institution{Arizona State University}
	\department{School of Electrical, Computer, and Energy Engineering}
	\city{Tempe}
	\state{AZ}
	\postcode{85287}
	\country{USA}
}

\author{Umit Y. Ogras}
\email{uogras@wisc.edu}
\affiliation{%
	\institution{University of Wisconsin-Madison}
	\department{Department of Electrical and Computer Engineering}
	\city{Madison}
	\state{WI}
	\postcode{53706}
	\country{USA}
}

\author{Yu Cao}
\email{Yu.Cao@asu.edu}
\orcid{0000-0003-1085-2189}
\affiliation{%
	\institution{Arizona State University}
	\department{School of Electrical, Computer, and Energy Engineering}
	\city{Tempe}
	\state{AZ}
	\postcode{85287}
	\country{USA}
}

\renewcommand{\shortauthors}{G. Krishnan, et al.}

\begin{abstract}

In-memory computing (IMC) on a monolithic chip \BT{for deep learning} faces dramatic \BT{challenges on} area, yield, and on-chip interconnection cost 
due to the ever-increasing model sizes. 2.5D integration or chiplet-based architectures interconnect multiple small chips (i.e., chiplets) to form a large computing system, presenting a feasible solution beyond a monolithic IMC architecture to accelerate large deep learning models. This paper presents a new benchmarking simulator, SIAM, to evaluate the performance of chiplet-based IMC architectures and explore the potential of such a paradigm shift \NC{in IMC architecture design}. SIAM integrates device, circuit, architecture, network-on-chip (NoC), network-on-package (NoP), and DRAM access \BT{models to realize} an end-to-end system. \NC{SIAM} is scalable \BT{in its support} of a wide range of deep neural networks (DNNs), customizable to various \BT{network} structures and configurations, and \BT{capable of} efficient design space exploration. We demonstrate the flexibility, scalability, and simulation speed of SIAM by benchmarking different state-of-the-art DNNs with CIFAR-10, CIFAR-100, and ImageNet datasets. We further calibrate the simulation results 
with a published silicon result, SIMBA. 
The chiplet-based IMC architecture obtained through SIAM shows
130$\times$ and 72$\times$ improvement in energy-efficiency for ResNet-50 on the ImageNet dataset compared to Nvidia V100 and T4 GPUs.

\end{abstract}



\maketitle

\section {Introduction} 


State-of-the-art deep neural networks (DNNs) have become more complex with wider, deeper, and more branched structures to cater to the needs of various applications~\cite{huang2017densely, howard2019searching}.
\BT{For instance,}
network architecture search (NAS) methods generate highly branched and complex DNNs, which increase compute and memory requirements~\cite{xie2019exploring, zoph2016neural}.
In-memory computing (IMC)-based architectures \BT{can support these network models} because of their ability to \BT{embed deep learning operations in the memory array, achieving massively parallel computing with high storage density.} 
Prior studies \BT{demonstrated} 
that crossbar-based IMC architectures \BT{with RRAM or SRAM significantly improved}
throughput and energy-efficiency for DNN accelerators~\cite{krishnan2020interconnect, mandal2020latency, shafiee2016isaac, imani2019floatpim, song2017pipelayer}. \BT{Such architectures usually assume all DNN weights are stored on a monolithic chip to minimize DRAM access and maximize parallel IMC computing.} However, \BT{as the DNN model size becomes larger and larger, this approach leads to increased chip area and on-chip} \BT{memory}.

\begin{figure}
     \centering
     \begin{subfigure}[b]{0.5\textwidth}
         \centering
         \includegraphics[width=\textwidth]{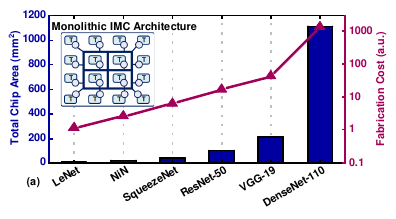}
     \end{subfigure}
     \hfill
     \begin{subfigure}[b]{0.48\textwidth}
         \centering
         \includegraphics[width=\textwidth]{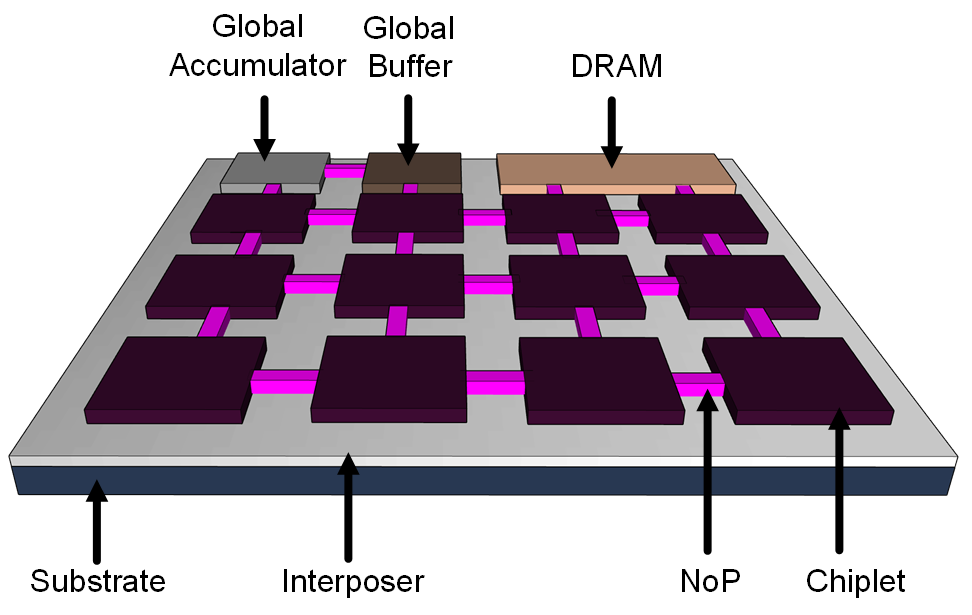}
     \end{subfigure}
     \vspace{-3mm}
     \caption{(a) Total chip area and fabrication cost for a monolithic RRAM-based IMC architecture for different DNNs~\cite{shafiee2016isaac}. Fabrication cost increases exponentially with an increase in total chip area (Appendix A details the method to calculate the fabrication cost). (b) 3-dimensional diagram showing the chiplet-based IMC architecture. The architecture includes an array of IMC chiplets, global buffer, global accumulator, and DRAM connected by an NoP. The figure is for representational purposes and not drawn to scale.}
     \label{fig:first_fig}
     \vspace{-4mm}
\end{figure}

Figure~\ref{fig:first_fig}(a) shows the total chip area
for a monolithic RRAM-based IMC architecture across different DNNs~\cite{krishnan2020interconnect}. 
Larger and branched DNNs like DenseNet-110~\cite{huang2017densely} result in a chip area of up to $1,200 \text{mm}^2$.
\rev{The increased area is attributed to the larger model size and the branched structure in DNNs. For example, ResNet-50 has 23M parameters and residual connections (branched connections). For an 8-bit precision, the mapping scheme in~\cite{shafiee2016isaac}, and a crossbar size of 128x128, results in 802 tiles. Here each tile consists of 16 IMC crossbar arrays. In addition, the presence of the residual connections results in increased buffer cost due to the need to store the activations of previous layers to perform residual addition operations in the ResNet DNN. For the same hardware configuration, LeNet-5 with 0.43M parameters requires 43 tiles, while DenseNet-110 with 28.1M parameters requires 2184 tiles for the same hardware configuration. 
Hence, the DNN size and structure influence the overall area and, in turn, fabrication cost.
}
Furthermore, higher chip area \BT{further causes}
lower yield and higher defects across the wafer~\cite{kannan2015enabling}\BT{,}
resulting in wasted area and 
a higher fabrication cost.
Figure
~\ref{fig:first_fig}(a) also shows the fabrication cost of the monolithic RRAM-based IMC architecture for different DNNs.
\NC{We see that} the fabrication cost increases exponentially with increased chip area (note that the cost is shown in \BT{the} logarithmic scale), thus making the
monolithic IMC architecture \BT{much} less cost-efficient \BT{if all weights are stored on a single chip}. 
Hence, there is an urgent need to address the increased 
fabrication cost of
IMC-based DNN accelerators~\cite{krishnan2020interconnect, song2017pipelayer, shafiee2016isaac}.

2.5D integration or chiplet-based architectures~\cite{shao2019simba, beck2018zeppelin, zimmer20190, lin20207} provide a promising alternative to monolithic
hardware architectures.
\BT{They}
integrate multiple chiplets 
through silicon interposers or organic substrates~\cite{turner2018ground, greenhill20173}. 
\BT{Compared to the monolithic chip,} the smaller chiplet size 
\BT{helps improve the design effort, yield, reduces defect ratio, and reduces fabrication cost.}
Figure~\ref{fig:first_fig}(b) shows a representative 3-dimensional diagram of a chiplet-based IMC architecture.
\NC{Chiplets comprise of memory units, computation blocks, and DRAM allowing for large-scale system integration.}
The interposer acts as an additional routing layer (network-on-package or NoP) that utilizes package-level signaling to connect different chiplets. 
Recent advances in package-level signaling have enabled a 2$\times$ improvement in bandwidth over board-level interconnections with 8$\times$ lower energy-per-bit~\cite{turner2018ground, lin20207}.

In a chiplet-based architecture, the design space \BT{parameters} primarily include IMC crossbar size, \BT{the} number of crossbars per chiplet, \BT{the} number of chiplets, network-on-chip (NoC), NoP, and the DNN \BT{structure}. \BT{These parameters need to be optimized systematically in order to exploit the potential provided by this new architecture.}
For example, SIMBA~\cite{shao2019simba}, a chiplet-based accelerator developed by Nvidia, utilizes 36 chiplets, with each chiplet consisting of 16 processing elements (PEs)~\cite{shao2019simba}.
While the underlying architecture ensures correct \BT{general-purpose} functionality, it does not guarantee optimal performance \BT{for various DNN applications.}
\BT{Therefore}, an extensive design space 
exploration is required to identify the optimal chiplet-based IMC architecture for DNN inference. 

In this work, we propose a novel chiplet-based IMC architecture simulator, SIAM, that integrates device, circuits, architecture, NoC, NoP, and DRAM access estimation under a single roof for
design space exploration.
We plan to open-source SIAM upon acceptance of this work. 
To the best of our knowledge, SIAM will be the first open-sourced chiplet-based IMC architecture simulator \BT{to promote architectural research in this emerging domain}.
SIAM includes four main components: partition and mapping engine, circuit and NoC engine, NoP engine, and DRAM engine.
\NC{A Python wrapper is used to interface each engine with one another.}
The wrapper also interfaces SIAM with popular deep learning frameworks such as TensorFlow and PyTorch.

SIAM provides a scalable solution that utilizes model-based and cycle-accurate simulation components, allowing for performance evaluation of a wide range of DNNs across multiple datasets.
It has a flexible architecture to support multiple DNN to IMC chiplet and crossbar partition and mapping schemes,
thus generating different types of chiplet-based IMC architectures.
\NC{Thus, SIAM provides a platform to enable comparisons across different chiplet-based IMC architectures and also between chiplet-based and monolithic IMC architectures.}
Furthermore, SIAM has a low simulation time \BT{to support the}
fast design and benchmarking
exploration.
For example, ResNet-110 with 1.7M parameters takes 12 minutes, while VGG-16 with 138M parameters takes 4.26 hours 
for benchmarking.

We demonstrate SIAM's capabilities by \BT{conducting experiments on} 
state-of-the-art DNNs such as ResNet-110~\cite{he2016deep} for CIFAR-10, VGG-19~\cite{simonyan2014very} for CIFAR-100, and ResNet-50~\cite{he2016deep} and VGG-16~\cite{simonyan2014very} for ImageNet datasets.
Furthermore, 
to \YC{evaluate SIAM at the system level}
, we calibrate SIAM against a published silicon result,
SIMBA~\cite{shao2019simba}\YC{, especially the scaling trend with the number of chiplets}.
\rev{The major contributions of this work are three-fold:}
\begin{itemize}
    \item \rev{We propose a complete framework, SIAM, that combines IMC circuit, NoC, NoP, and DRAM performance evaluation under a single roof. \textit{SIAM is the first simulator to provide support for hardware performance evaluation of chiplet-based IMC architectures.} We carefully model the components of architecture like the IMC circuit, NoC, network-on-package (NoP), and DRAM.}
    \item \rev{We provide a high degree of freedom to the user through different mapping schemes and customizable architectural parameters for IMC circuit, NoC, NoP, and DRAM components. We demonstrate different architectural design space exploration experiments that can be performed using SIAM.}
    \item \rev{Extensive experimental evaluation of the SIAM simulator for different DNNs across CIFAR-10, CIFAR-100, and ImageNet datasets. Furthermore, we perform detailed experiments to calibrate the simulator to a real-design, SIMBA~\cite{shao2019simba}, making SIAM a reliable and accurate simulator.
    For ResNet-50 on ImageNet, the generated chiplet-based IMC architecture achieves 130$\times$ and 72$\times$ improvement in energy-efficiency compared to Nvidia V100 and T4 GPUs.}
\end{itemize}
\section{Related Work}


\subsection{In-Memory Computing}

In-memory computing \BT{is} 
a promising alternative to conventional von-Neumann architectures~\cite{shafiee2016isaac, imani2019floatpim, yin2019vesti}.
Prior work\BT{s mostly focus} on a monolithic IMC-based architecture with all weights on-chip \BT{to minimize the access to external memory}.
Such an assumption leads to a higher area cost and, in turn, a higher fabrication cost, as shown in Figure~\ref{fig:first_fig}(a).
Higher area cost leads to lower yield and \BT{other difficulties in fabrication.}
To address this, in this work, we develop a benchmarking tool, SIAM, that utilizes a chiplet-based IMC architecture \NC{for} DNN inference.
Each chiplet \BT{integrates} 
an array of IMC crossbar\BT{s} and associated peripheral circuits 
to perform the MAC operations.
\NC{The chiplets also consist of buffers, accumulator circuits, pooling units, non-linear activation units, NoP driver circuit, and NoP router all interconnected using an NoC~\cite{marculescu2008outstanding}.} 
\BT{It} \BT{consumes} 
less per-chip area than a monolithic IMC architecture and results in lower fabrication cost and higher yield while providing similar system-level inference performance as the monolithic IMC architecture.

\subsection{Chiplet-based Architectures}
2.5D or chiplet-based architectures utilize package-level integration of small blocks, known as chiplets.
Chiplet-based architectures provide a promising solution to build cost-effective large-scale systems for complex applications such as DNN inference.
\NC{Chiplets consist of computation and memory units that form the primary building blocks of any architecture.}
Multiple chiplets are connected using an on-package network integrated into a silicon interposer or an organic substrate. 
\rev{Prior work has shown multiple chiplet-based architectures for various applications such as DNN acceleration, general purpose system-on-chips (SoCs), and recommendation systems~\cite{yin2018modular, shao2019simba, beck2018zeppelin, erett2018126mw, zimmer20190, lin20207, hwang2020centaur},
which present different on-package signaling techniques and associated circuitries.
We limit the description of these works to three for brevity.
A fine-grained chiplet architecture for deep learning inference is proposed in~\cite{shao2019simba}.
A custom NoP driver and associated interconnect that provides improved performance for large-scale DNN inference is utilized. A high performance SoC with up to 8 CPU cores that can scale across different market segments (edge and cloud) is proposed in~\cite{beck2018zeppelin}. 
A custom infinity fabric (NoP) that utilizes a physical layer for in-die and across package communication is utilized. Finally,~\cite{hwang2020centaur} proposed a chiplet-based hybrid sparse-dense accelerator that utilizes a package-integrated CPU+FPGA system for personalized recommendation applications. A sparse accelerator is used for high-throughput embedding gather and reduction operations, while a dense accelerator is used for accelerating the compute-intensive DNN layers.
}
However, all 
these prior studies focus on custom designs for specific applications, leaving \BT{little} 
room for
architecture-level benchmarking \BT{and} design space exploration for chiplet-based architectures.
Furthermore, they focus on CMOS-based conventional von-Neumann architectures, with little or no focus on IMC architectures. 
To address this, SIAM provides a benchmarking tool for IMC architectures based on chiplet structures \BT{with design space exploration capabilities}.


\subsection{Benchmarking Tools}
\begin{table}[t]
\caption{\rev{Comparison between different IMC Simulators}}
\label{tab:simulator_comparison}
\resizebox{0.9\textwidth}{!}{%
\begin{tabular}{@{}c|c|c|c|c|c@{}}
\toprule
\textbf{Simulator} &
  \textbf{Architecture} &
  \textbf{Circuit} &
  \textbf{Interconnect} &
  \textbf{NoP Interconnect} &
  \textbf{DRAM} \\ \midrule
GenieX~\cite{chakraborty2020geniex} & Monolithic & SPICE-based     & No & No & No \\ \midrule
RxNN~\cite{jain2020rxnn}    & Monolithic  & SPICE-based & No                & No & No \\ \midrule
NeuroSim~\cite{peng2019dnn+}   & Monolithic & SPICE-based     & P2P (H-Tree)                       & No & No \\ \midrule
MNSIM~\cite{zhu2020mnsim}     & Monolithic & Behavior model  & NoC-mesh                       & No & No \\ \midrule
\textbf{SIAM} &
  \textbf{Monolithic \& Chiplet} &
  \textbf{\begin{tabular}[c]{@{}c@{}}SPICE and \\ Behavioral Model\end{tabular}} &
  \textbf{\begin{tabular}[c]{@{}c@{}}NoC-mesh, NoC-tree, \\ and H-Tree\end{tabular}} &
  \textbf{\begin{tabular}[c]{@{}c@{}}Supported \\ (driver and interconnect)\end{tabular}} &
  \textbf{Supported} \\ \bottomrule
\end{tabular}
}
\end{table}
Benchmarking tools enable a fast 
design space exploration. 
Prior work have proposed a number of benchmarking tools for IMC-based architectures\rev{~\cite{peng2019dnn+, zhu2020mnsim, chakraborty2020geniex, jain2020rxnn, banagozar2019cim, russo2021lambda, krishnan2021interconnect}} and conventional von-Neumann architectures~\cite{samajdar2018scale, binkert2011gem5, qureshi2019gem5}. 
Authors in~\cite{peng2019dnn+} provide a DNN inference benchmarking tool for different device technologies.
%
%
The architecture utilizes an array of IMC crossbars and peripheral circuits connected by point-to-point (P2P) interconnect for on-chip data movement.
Recently, authors in~\cite{zhu2020mnsim} proposed a behavioral performance benchmarking tool for IMC architectures.
They utilize individual models for each component in the IMC architecture to perform the estimation. 
All the prior simulators suffer from two significant drawbacks.
\textit{First}, they assume a custom monolithic IMC architecture for a given DNN. 
With a custom design, the architecture \BT{and resource utilization are different} 
for each DNN, i.e., for each network, a specific IMC architecture is generated and evaluated to provide the desired inference hardware performance.
\NC{Such an assumption leads to a need for designing an IMC architecture that is specific to a DNN.}
\NC{\textit{Second,} the authors provide support only for monolithic IMC architectures, with no support for chiplet-based IMC structures}.

\BT{In contrast to all this research,} this \BT{work} proposes a novel\BT{, general-purpose} chiplet-based IMC architecture simulator, SIAM, that integrates device, circuits, architecture, NoC, NoP, and DRAM access estimation for design space exploration.
SIAM utilizes smaller individual chiplets connected by the NoP fabric. 
Smaller chiplets lead to a lower area and fabrication cost. 
Furthermore, SIAM provides flexibility for both custom and homogeneous (generic) IMC architectures, allowing a higher degree of freedom for the user.
\rev{Table~\ref{tab:simulator_comparison} shows the comparison between different popular IMC simulators. GenieX~\cite{chakraborty2020geniex} and RxNN~\cite{jain2020rxnn} target monolithic IMC architectures. The performance of the computing fabric is obtained through SPICE-based models. Neither of these simulators estimate the performance of the communication fabric. NeuroSim~\cite{peng2019dnn+} and MNSIM~\cite{zhu2020mnsim} utilize an H-tree based point-to-point (P2P) network and mesh-NoC, respectively as communication fabric. Both these simulators target monolithic IMC architectures. In contrast to all these simulators, SIAM provides the first chiplet-based IMC simulator that supports circuit, interconnect (NoC and point-to-point), DRAM cost estimation, and NoP interconnect evaluation.
}

\section{Chiplet-based IMC Architecture}
%
\begin{figure}[t]
	\centering
	\includegraphics[width=1\textwidth]{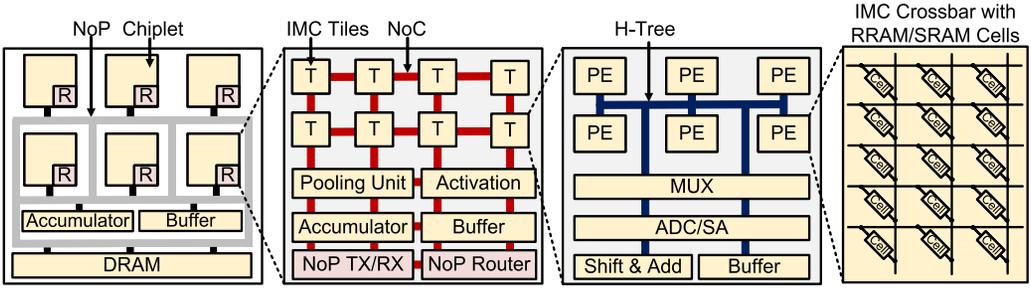}
	\vspace{-4mm}
	\caption{Chiplet-based IMC architecture utilized within SIAM. The architecture consists of IMC chiplets, global accumulator, buffer, and DRAM connected using an NoP. SIAM supports both SRAM- and RRAM-based IMC architectures. \BT{An NoP is applied for inter-chiplet communication and} an NoC is utilized within the chiplet for intra-chiplet communication.}
	\label{fig:arch_chiplet}
	\vspace{-4mm}
\end{figure}
%

\textbf{Overview:} This section presents the underlying chiplet-based IMC architectures supported by SIAM, \BT{for} \textit{homogeneous (generic)} and \textit{\BT{custom design}\BT{s}}.
In a homogeneous architecture,
the number of chiplets is fixed and is determined by the user.
A custom architecture consists of the required number of chiplets to map the DNN under consideration.
In both cases, the chiplet structure has \BT{a} fixed number of IMC crossbar arrays inside (user-defined).

Figure~\ref{fig:arch_chiplet} shows a homogeneous chiplet-based IMC architecture utilized by SIAM.
The entire architecture consists of an array of chiplets that include IMC compute units, a global accumulator, a global buffer, and a DRAM chiplet.
The chiplets are connected using an NoP fabric.
The global accumulator and buffers are used to perform the accumulation across chiplets, and the DRAM chiplet is used to store the pre-trained DNN weights. 
In this work, we assume that all weights are transferred to the IMC chiplets from the DRAM chiplet before performing the DNN inference, consistent with prior works~\cite{krishnan2020interconnect, shafiee2016isaac, imani2019floatpim}.

\textbf{Intra-Chiplet IMC Architecture:} Each IMC chiplet consists of an array of IMC tiles connected using an NoC.
The IMC tiles consist of an array of processing elements (PEs) or crossbar arrays. 
SIAM \BT{currently} supports both RRAM- and SRAM-based IMC crossbar architectures.
The IMC crossbars utilize analog computation to perform the multiply-and-accumulate (MAC) operation.
Each IMC crossbar has associated peripheral circuitry (e.g. column multiplexers, \BT{analog-to-digital converter (ADC)}, shift and add circuits, etc.).
A column multiplexer is used to share a flash 
ADC or sense amplifier (SA) across multiple columns of the IMC crossbar.
The ADC converts the analog output from the IMC crossbar to the digital domain. 
Next, the ADC output is accumulated based on the bit significance using shifter and adder circuits to extract the computed MAC output. 
Finally, the overall result is generated by accumulating the outputs from each IMC crossbar across the entire input.
\BT{Note that our} architecture does not use a digital-to-analog converter (DAC), 
\BT{and it instead} employs sequential \BT{bit-serial computing} 
for multi-bit inputs.
Furthermore, each chiplet consists of a pooling and activation unit. 
The pooling unit supports both max and average pooling operations, while \BT{the} activation \BT{unit} supports ReLU and sigmoid functions.

\textbf{Interconnect:} The IMC chiplets are connected at the tile-level (within chiplet) using an NoC.
A point-to-point (P2P) interconnect, such as H-Tree, is used for communication at the PE-level.
Each tile within the chiplet has a five-port router that performs the data scheduling and X--Y routing through the NoC.
The NoC can be configured for multiple flit width
and operating frequencies \BT{by the user.}
The array of chiplets are interconnected using an NoP that utilizes the interposer for routing.
A passive interposer is implemented within SIAM where the interposer does not contain any active elements like repeaters.
Each chiplet consists of a dedicated NoP transmitter and receiver (TX/RX) circuit and an NoP router.
The router performs the packet scheduling and utilizes a dedicated port to transmit data to the TX/RX circuit.
The custom TX/RX circuit can be configured for a given signaling technique to achieve data transfer across the NoP~\cite{turner2018ground, lin20207}.
The architecture also includes a clocking circuit (e.g.: LC-PLL~\cite{poulton20130}) for the TX/RX circuit. 
The NoP interconnect properties such as wire resistance, capacitance, and inductance are carefully modeled by utilizing the PTM models~\cite{sinha2012exploring} following prior works~\cite{turner2018ground, lin20207}. 
Section~\ref{sec:nop_engine} details the modeling of both the NoP interconnect and the driver.
\section{SIAM Simulator}
\begin{figure}[t]
	\centering
	\includegraphics[width=1\textwidth]{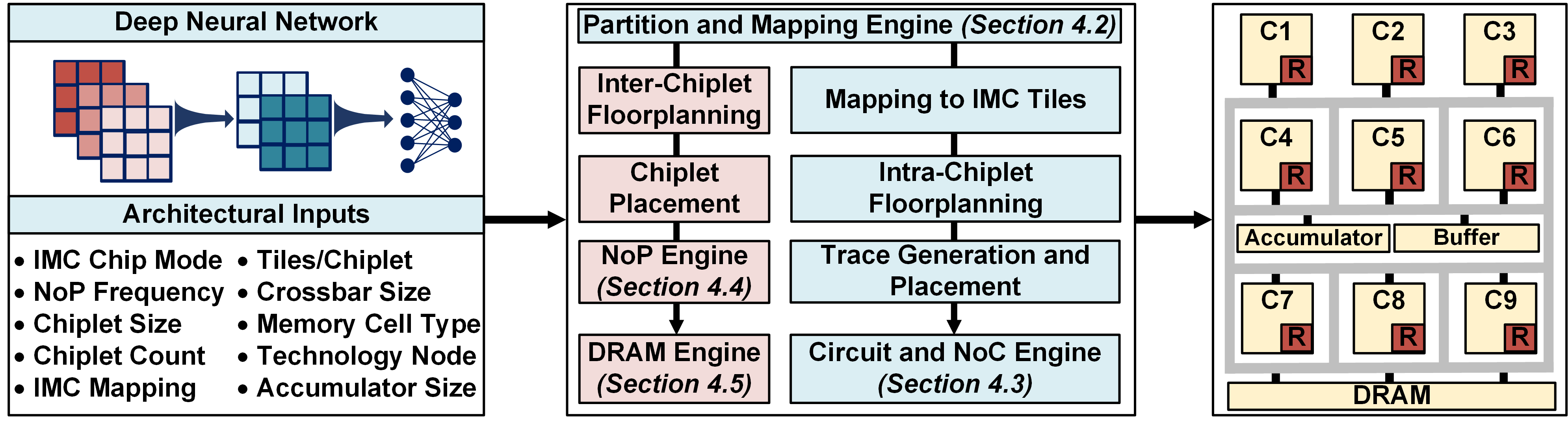}
	\vspace{-8mm}
	\caption{\rev{An overview of the proposed chiplet-based IMC benchmarking simulator, SIAM. SIAM incorporates device, circuits, architecture, NoC~\cite{jiang2013detailed}, NoP, and DRAM access model~\cite{kim2015ramulator} under a single roof for system-level analysis of chiplet-based IMC architectures.}}
	\label{fig:overview}
\end{figure}
%

\subsection{Overview}
SIAM provides a unified framework for performance benchmarking of chiplet-based IMC architectures, as shown in Figure~\ref{fig:overview}. 
%
SIAM operates on user inputs to generate the chiplet-based IMC architecture and \BT{to benchmark} the \BT{corresponding} 
hardware performance.
The hardware performance metrics include area, energy, latency, energy efficiency, power, leakage energy, and IMC utilization.
The overall simulator is developed using Python and C++ programming languages.
A top-level Python wrapper is built to combine the different components within the simulator. 
Furthermore, SIAM interfaces with popular deep learning frameworks such as PyTorch and Tensorflow. 
Thus, SIAM supports multiple network structures in current literature (as shown in Section~\ref{sec:expt}) and can be used for exploring NAS techniques.
Table~\ref{tab:main_input} shows the user inputs and associated descriptions of the SIAM benchmarking tool. %

\begin{table}[t]
\vspace{-3mm}
\caption{Definition of the User Inputs to SIAM}
\label{tab:main_input}
\resizebox{\textwidth}{!}{%
\begin{tabular}{@{}l|l|ll@{}}
\toprule
\textbf{User Input}  & \textbf{Description}              & \multicolumn{1}{c|}{\textbf{User Input}}     & \textbf{Description}                          \\ \midrule
\multicolumn{2}{c|}{\textbf{DNN Algorithm}}                  & \multicolumn{2}{c}{\textbf{Device and Technology}}                                           \\ \midrule
Network Structure    & DNN network structure information  & \multicolumn{1}{l|}{Tech Node}               &  Technology node for fabrication               \\
Data Precision  & Weights and activation precision   & \multicolumn{1}{l|}{Memory Cell}             & RRAM or SRAM                                  \\
Sparsity             & DNN layer-wise sparsity           & \multicolumn{1}{l|}{Bits/Cell}               & Number of levels in RRAM                      \\ \midrule
\multicolumn{2}{c|}{\textbf{Intra-Chiplet Architecture}} & \multicolumn{2}{c}{\textbf{Inter-Chiplet Architecture}}                                      \\ \midrule
Crossbar Size        & IMC crossbar array size           & \multicolumn{1}{l|}{Chip Mode}               & Monolithic or chiplet-based IMC architecture   \\
Buffer Type          & SRAM or Register File             & \multicolumn{1}{l|}{Chiplet Structure}       & Homogeneous or custom chiplet structure       \\
ADC Resolution       & Bit-precision of flash ADC        & \multicolumn{1}{l|}{Chiplet Size}            & Number of IMC tiles within each chiplet       \\
Read-out Method      & Sequential or Parallel            & \multicolumn{1}{l|}{Total Chiplet Count}     & Fixed count or DNN specific custom count      \\
NoC Topology         & Mesh or Tree                      & \multicolumn{1}{l|}{Global Accumulator Size} & Size of global accumulator                    \\
NoC Width            & Number of channels in the NoC     & \multicolumn{1}{l|}{NoP Frequency}           & Frequency of the NoP driver and  interconnect \\
Frequency            & Frequency of operation            & \multicolumn{1}{l|}{NoP Channel Width}       & Number of parallel links for TX and RX        \\ \bottomrule
\end{tabular}%
}
\vspace{-3mm}
\end{table}
%

SIAM consists of four engines\BT{:} 


\begin{itemize}
    \item \makebox[6cm]{Partition and mapping engine (Python)\hfill}\inlineitem Circuit and NoC engine (C++)
    \item \makebox[6cm]{NoP engine (Python and C++)\hfill}\inlineitem DRAM engine (Python and C++)
\end{itemize}

Each engine functions independently on a subset of the user inputs, while communicating with each other using the top-level Python wrapper.
To further understand the framework, we detail the simulation flow used \BT{for} 
SIAM 
in Figure~\ref{fig:overview}.
\textit{First}, SIAM takes the user inputs and performs the layer partition and mapping onto the chiplets and IMC crossbars using the partition and mapping engine.
The outputs include \BT{the} structure of IMC architecture, the number of chiplets and IMC tiles required per layer, utilization of the IMC architecture, intra-chiplet and inter-chiplet data movement volume, and \BT{the} number of global accumulator accesses.
Next, the intra-chiplet and global circuit performance are evaluated using the circuit and NoC engine. 
The engine provides the hardware performance metrics such as area, energy, and latency for the intra-chiplet and global circuit operations across all chiplets.
Simultaneously, the NoP engine evaluates the cost of the interconnect, router, and driver associated with the chiplet-to-chiplet data movement. 
Finally, the DRAM engine determines the cost of the memory accesses and provides the energy and latency performance metrics.
All engines except the partition and mapping engine work simultaneously, thus reducing the total simulation time.
We note that SIAM also supports benchmarking of conventional monolithic IMC architectures.
In the following sections\BT{,} we detail the four engines that \BT{represent} the core functionality within SIAM.

\subsection{Partition and Mapping Engine}\label{sec:part_and_mapping}

%
\begin{algorithm}[t]
\SetNoFillComment
\caption{Partition and Mapping of DNN layers} \label{algo:partition_mapping}
\textbf{Input:} DNN structure, chiplet count ($C$), chiplet size ($S$), the number of DNN layers ($N_l$) \\
\textbf{Output:} Layer-wise chiplet partition ($\mathcal{L} \rightarrow \mathcal{P}$) and layer to chiplet mapping ($\mathcal{L} \rightarrow \mathcal{C}$)\\
\SetAlgoLined
\For{$i = 1:N_{l}$}{
$N^{Chiplet} = 0,~N^{Chiplet}_i = 0,~N^{Total}_i = 0$  \hspace{4mm}   \tcc{\textbf{Initialize variables}}
\tcc{\textbf{Layer-wise Mapping ($\mathcal{L} \rightarrow \mathcal{C}$)}}
Calculate number of rows of IMC crossbars ($N^r_i$) to map layer $i$ (Equation~\ref{eq:Nr}) \\
Calculate number of columns of IMC crossbars ($N^c_i$) to map layer $i$ (Equation~\ref{eq:Nr}) \\
$N^{Total}_i$ = $N^r_i$ $\times$ $N^c_i$ \hspace{4mm} \tcc{\textbf{Total number of IMC crossbars for layer $i$}}
\tcc{\textbf{Layer-wise Partitioning ($\mathcal{L} \rightarrow \mathcal{P}$)}}
$N^{Chiplet}_i = \bigg\lceil \frac{N^{Total}_i}{S} \bigg\rceil$ \hspace{4mm} \tcc{\textbf{Calculate the number of chiplets for layer $i$}}
$N^{Chiplet} += N^{Chiplet}_i$ \hspace{4mm} \tcc{\textbf{Increment total chiplets in the architecture}}
\If{Homogeneous Mapping}{ 	
\If{$N^{Chiplet}$ > $C$}{ 	
   \textbf{exit()} \hspace{4mm} \tcc{\textbf{Error: Exceeded the maximum number of chiplets}}
}
}
\tcc{\textbf{Partition and Mapping completed for layer $i$}}
}
\end{algorithm}
%
Algorithm~\ref{algo:partition_mapping} describes the step-by-step operation of the partition and mapping engine. 
The engine performs the partition of \BT{DNN} layers to the IMC chiplets and the corresponding mapping to the IMC crossbar arrays.
The partition and mapping is performed layer-wise for the entire DNN.
The engine utilizes user inputs such as the DNN structure, DNN weight precision, IMC chiplet mapping scheme, size of the IMC chiplet, and the IMC crossbar size, among others.

We first discuss the IMC mapping scheme utilized in SIAM. 
For a given layer $i$, let the weight matrix be W$_i$ represented by Kx$_i$ $\times$ Ky$_i$ $\times$ Nif$_i$ $\times$ Nof$_i$, where Kx and Ky represent the kernel size, Nif the number of input features, and Nof the number of output features.
\BT{We adopt the following mapping scheme, similar to that in}
~\cite{krishnan2020interconnect, shafiee2016isaac}:
%
\begin{equation}\label{eq:Nr}
    N^r_i = \Big\lceil \frac{Kx_i \times Ky_i \times Nif_i}{(PE_x)} \Big\rceil;\hspace{7mm} N^c_i = \Big\lceil \frac{Nof_i \times N_{bits}}{(PE_y)} \Big\rceil
\end{equation}
%

In the above equation, N$^r_i$ and N$^c_i$ are the required number of rows and columns of IMC crossbars needed to map the layer $i$ of the DNN.
N$_{bits}$, PE$_x$ and PE$_y$ represent the DNN weight precision, the number of rows and columns in the IMC crossbar array, respectively.
The product of N$^r_i$ and N$^c_i$ is the total number of required IMC crossbar arrays N$^{Total}_i$ to map layer $i$ of the DNN (line 7 of Algorithm~\ref{algo:partition_mapping}).

\begin{figure}[t]
	\centering
	\includegraphics[width=0.8\textwidth]{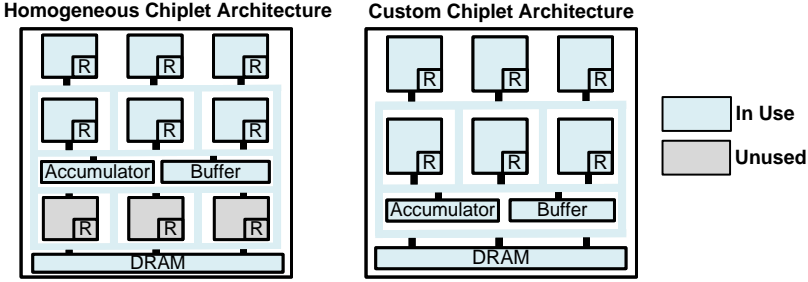}
	\vspace{-5mm}
	\caption{Representative figure showing the two generated chiplet-based IMC architectures \BT{for the same DNN}, homogeneous (left) and custom (right), from the supported partition schemes in SIAM. Homogeneous architecture is generic, while custom architecture is DNN specific. R refers to the NoP router.}
	\label{fig:mapping_arch}
	\vspace{-4mm}
\end{figure}

SIAM can generate \NC{(a)} \textit{homogeneous} and \NC{(b)} \textit{custom} chiplet-based IMC architectures using two types of chiplet partitions.
Figure~\ref{fig:mapping_arch} shows the two generated architectures based on the homogeneous and custom chiplet partitioning.
\rev{The partition and mapping engine assumes that DNN layers cannot be partitioned across multiple chiplets, and a single chiplet can support multiple layers to achieve high chiplet utilization (shown in Section~\ref{sec:expt}).
Since each layer of the DNN contains a large number of multi-bit weights, multiple chiplets consisting of IMC crossbar arrays are required to map the whole layer.
If one layer is distributed across chiplets, it increases the overhead in terms of the control logic for routing the inputs to the respective chiplets, an increased volume of inter-chiplet data communication, and a higher chiplet-to-chiplet communication energy and latency.}
During the partition of layers across multiple chiplets, the engine divides the layer uniformly across the chiplets, thus avoiding the workload imbalance issue.
Based on the total number of required IMC crossbar arrays, N$^{Total}_i$, the engine determines the number of chiplets necessary to map the layer $i$ of the DNN \BT{as:} \NC{$N^{Chiplet}_i = \bigg\lceil \frac{N^{Total}_i}{S} \bigg\rceil$}, 
%
where S denotes the total number of IMC crossbar arrays within a chiplet (size of the chiplet).

Next, the total number of chiplets in the architecture (N$^{Chiplet}$) is determined (line 9 of Algorithm~\ref{algo:partition_mapping}).
In the \textit{homogeneous chiplet partition} scheme, a fixed number of chiplets (user input) is used to map the DNN.
Hence, the engine compares the total number of chiplets in the architecture (N$^{Chiplet}$) with the maximum available chiplets ($C$).
If greater, then the engine throws an error and requests for an increase in the number of available chiplets in the architecture.
\BT{If} lesser, the engine continues the partition and mapping for the subsequent layers in the DNN.

In the \textit{custom partition scheme}, the architecture consists of the required number of chiplets to map the DNN.
Hence, there is no maximum limit in the number of available chiplets within the architecture.
Such a design results in a fully-custom architecture specific to the DNN under consideration.
Each chiplet has the same structure with a fixed number of IMC tiles, where each tile consists of IMC crossbar arrays and associated peripheral circuitry.
Thus, SIAM provides a platform to perform comparison between homogeneous (generic) and custom-designed chiplet-based IMC architectures.

\NC{After partitioning and mapping layers} onto the IMC chiplets, the engine determines the total volume of data communicated within the chiplet and across chiplets.
Simultaneously, when a layer is partitioned across chiplets, the global accumulator is used to generate the overall layer output. 
The engine determines the number of additions performed by the global accumulator and the number of global buffer accesses.
Overall, the engine provides the layer partition across chiplets, the number of required chiplets and IMC crossbars, IMC crossbar utilization, volume of intra- and inter-chiplet data movement, and the number of the global accumulator and buffer accesses.
The other engines (circuit, NoC and NoP) then utilize these outputs to evaluate the hardware performance of the chiplet-based IMC architecture.

\subsection{Circuit and NoC Engine}
\begin{figure}[t]
	\centering
	\includegraphics[width=0.9\textwidth]{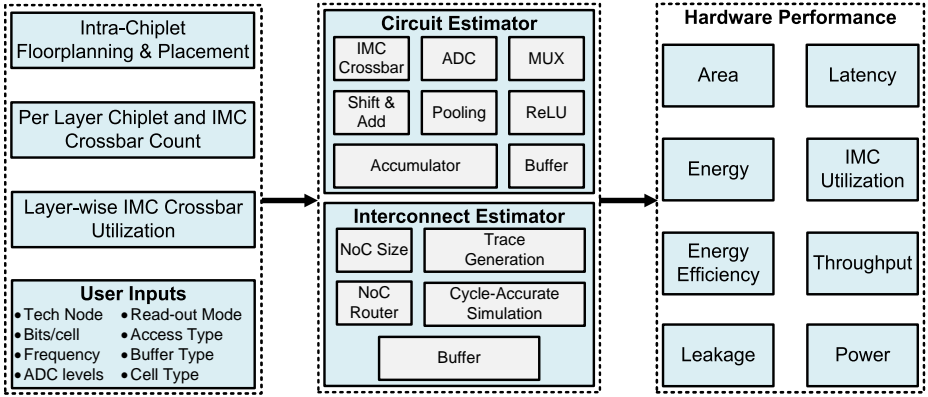}
	\vspace{-3mm}
	\caption{Block diagram of the Circuit and NoC engine within SIAM. The engine utilizes a separate circuit and NoC simulators that perform the overall hardware performance estimation.}
	\label{fig:ckt_noc}
	\vspace{-2mm}
\end{figure}

After completing the partition and mapping of the DNN, SIAM performs the inter- and intra-chiplet floorplanning and placement, thus determining the entire chiplet-based IMC architecture.
Thereafter, the circuit and NoC engine estimates the hardware performance.
Figure~\ref{fig:ckt_noc} shows the block diagram of the circuit and NoC engine.
The engine \BT{employs} 
a model-based estimator for the circuit part, \BT{and} 
a trace-based estimator for the interconnect part.

\subsubsection{Circuit Estimator}
The circuit estimator evaluates the overall hardware performance of each chiplet, global accumulator, and global buffer in the overall architecture.
The inputs to the engine include the intra- and inter-chiplet placement, per layer chiplet and IMC crossbar count, layer-wise IMC utilization, technology node, frequency of operation, IMC cell type, \BT{the} number of bits per cell, read-out mode (row-by-row or parallel), and ADC precision, among others.
The intra-chiplet circuits include the IMC crossbar array and associated peripheral circuits, buffer, accumulator, activation unit, \BT{and} the pooling unit. 
The peripheral circuits include the ADC, multiplexer circuit, shift and add circuit, and decoders.
We calibrate the circuit estimator with NeuroSim~\cite{peng2019dnn+}.
%

The circuit estimator evaluates the performance of the entire chiplet-based IMC architecture in a layer-wise manner.
Each chiplet performs the computations of a subset of layers in the DNN.
For a given DNN layer $i$, the \BT{chiplet count} per layer, \BT{the IMC crossbar count} per layer, and \BT{the} IMC utilization \BT{values} are taken from the partition and mapping engine.
Area, energy, and latency are estimated in a bottom-up manner, i.e., the estimation starts from the device level and moves up to the circuit level and, finally, the architecture level. 
Based on user input\BT{s} such as technology node, IMC cell type, IMC crossbar size, IMC utilization, ADC precision, and read-out mode, the estimator evaluates the cost of a single crossbar and associated peripheral circuits. 
The estimator repeats the process for all IMC crossbars within the chiplet for the given layer $i$ in the DNN.
Next, the estimator evaluates the buffer cost, shift and adder circuitry, and the accumulator within the chiplet.
After that, the pooling and activation units are evaluated to \BT{obtain} 
the total area, energy, and latency of the IMC chiplet.
At the chiplet-level, the global accumulator and global buffer accumulate the partial sum of a layer across chiplets.
The circuit estimator utilizes the number of additions performed, the data volume from each chiplet, and the accumulator size (user input) to determine the area, energy, and latency of the global accumulator and buffer.
Finally, based on the number of chiplets required for layer $i$ of the DNN, the circuit estimator repeats the estimation for all chiplets to determine the overall hardware performance.

\subsubsection{NoC Estimator}
\begin{algorithm}[t]
\SetNoFillComment
\caption{NoC (or NoP) Trace Generation} \label{algo:noc_trace}
\textbf{Input:} Number of tiles for each layer (for each chiplet in case of NoP) ($|\mathcal{T}|$), Number of input activations for each layer($A$),  Number of chiplets ($\mathcal{C}$), Layer to chiplet mapping ($\mathcal{L} \rightarrow \mathcal{C}$), Quantization bit ($Q$), Bus width ($W$)\\
\textbf{Output:} Trace file for each chiplet ($tr^c$) \\
\SetAlgoLined

\For {$c = 1:|\mathcal{C}|$} {

Find index of the first layer ($L_c^S$) and the last layer ($L_c^E$) in the chiplet from $\mathcal{L} \rightarrow \mathcal{C}$ \\

\For{$l = L_c^S:L_c^E$}{

$k = 0$ \hspace{4mm}\tcc{\textbf{Initialize timestamp}}

Find index of first source tile ($T_l^S$) and last source tile ($T_l^E$) \\
Find index of first destination tile ($T_{l+}^S$) and last destination tile ($T_{l+}^E$) \\

$N_{p} = \lceil \frac{A(l)Q}{W} \rceil $\hspace{4mm} \tcc{\textbf{Number of packets}}

\For{$n = 1:N_{p}$}{

\For{$s = T_l^S:T_l^E$}{

\For{$d = T_{l+}^S:T_{l+}^E$}{

$tr^c \leftarrow [tr^c; [s, d, k]]$ \\

$k \leftarrow k+1$ \hspace{4mm}\tcc{\textbf{Increment timestamp}}

}

$k \leftarrow k+1$ \hspace{4mm}\tcc{\textbf{Increment timestamp}}

}
}

}

}
\end{algorithm}

\rev{Communication plays a crucial role in the hardware performance of DNN accelerators~\cite{mandal2020latency}.
A detailed description of communication-centric DNN accelerators can be found in~\cite{nabavinejad2020overview}.}
Each layer within the DNN sends a significant amount of data to other layers. 
Authors in~\cite{krishnan2020interconnect} show that communication alone incurs up to 90\% of the total inference latency for DNNs.
Therefore, designing an efficient communication protocol for DNNs is of supreme importance.
Hence, we carefully incorporate the cost of communication between multiple layers within a chiplet.
We consider an NoC for intra-chiplet communication since NoC is the standard interconnect fabric used in the SoC-domain~\cite{jeffers2016intel, mandal2021energy}.
We customize a cycle-accurate NoC simulator, BookSim~\cite{jiang2013detailed}\BT{,} to evaluate the NoC performance.
First, a trace file is generated for each chiplet following Algorithm~\ref{algo:noc_trace}.
The algorithm takes the number of tiles for each layer, \BT{the} number of input activations for each layer, number of chiplets, layer to chiplet mapping, quantization bit\BT{-precision}, and bus width.
From these inputs, we find the indices of the first and the last layer of each chiplet.
Next, for each layer in each chiplet, we find the source and destination tile information as shown in lines 7--8 of Algorithm~\ref{algo:noc_trace}.
The number of packets for each source-destination pair is then calculated.
After that we iterate over the number of packets, the number of source tiles, and the number of destination tiles to obtain a trace in the form of a tuple consisting of the source tile \BT{ID}, destination tile \BT{ID}, and the timestamp.
The timestamp variable is reset to zero after generating trace for each pair of layers.
Then, the trace file is simulated using BookSim to obtain the area, energy, and latency for on-chip communication within each chiplet.

\subsection{NoP Engine}\label{sec:nop_engine}

\begin{figure}[t]
     \centering
     \begin{subfigure}[b]{0.48\textwidth}
         \centering
         \includegraphics[width=\textwidth]{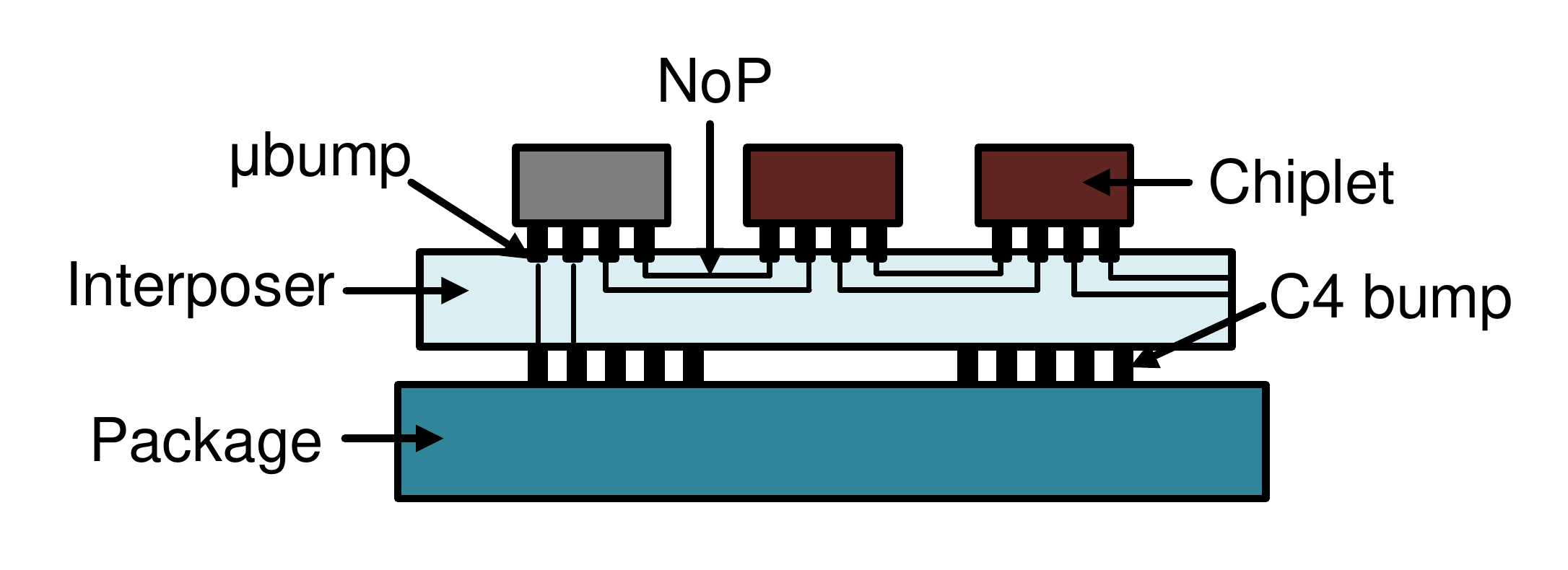}
     \end{subfigure}
     \hfill
     \begin{subfigure}[b]{0.45\textwidth}
         \centering
         \includegraphics[width=\textwidth]{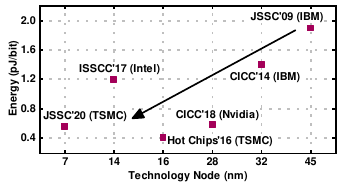}
     \end{subfigure}
     \caption{(Left) Cross-sectional image of the NoP interconnect. The NoP is routed within the interposer connecting different chiplets across the architecture. $\mu$bumps connect the chiplets to the interposer, (Right) Energy per bit for different NoP driver circuit and signaling techniques proposed in prior works.}
     \label{fig:crosssectional}
     \vspace{-4mm}
\end{figure}

The NoP connects different chiplets through a silicon interposer or organic substrate.
\NC{It performs the on-chip data movement using special signaling techniques and driver circuits, as shown in~\cite{poulton20130, lin20207}.}
Figure~\ref{fig:crosssectional} (left) shows the cross-sectional image of a 2.5D integration with chiplets and an interposer.
Modeling the NoP performance has many challenges due to the complex interconnect structure, specialized driver architectures, and the corresponding signaling techniques.

To this end, \BT{our} 
NoP engine models each component in the NoP for accurate performance estimation.
Figure~\ref{fig:crosssectional} (right) shows various NoP implementations with the corresponding energy-per-bit (E$_{bit}$) proposed in prior works.
\rev{There are two main components of the NoP performance evaluation: 1) NoP latency estimation and 2) NoP area and power estimation. }

\rev{\noindent\textbf{NoP latency estimation:} The engine utilizes a cycle-accurate simulator to perform the interconnect evaluation.}
First, based on the chiplet-to-chiplet data volume generated by the partition and mapping engine, the NoP engine utilizes  Algorithm~\ref{algo:noc_trace} (same as the algorithm for NoC) to generate the trace for the NoP.
These traces are simulated using a cycle-accurate simulator or the NoP estimator (a customized version of BookSim to incorporate a trace-based simulation) to obtain the latency of the NoP interconnect.

\rev{\noindent\textbf{NoP area and power estimation:} To estimate the area and power consumption of the NoP, we first obtain the interconnect parameters for the NoP, which include wire length, pitch, width, and stack-up.}
\NC{We use these parameters to determine the interconnect capacitance and resistance using the PTM interconnect models~\cite{sinha2012exploring}.}
Next, based on the capacitance and resistance\BT{,} the timing parameters for the interconnect are generated and compared with the target bandwidth.
\BT{If the timing parameters do not satisfy the bandwidth}, 
the NoP engine chooses the maximum allowable bandwidth.

%
\begin{algorithm}[t]
\SetNoFillComment
\caption{Computation of NoP driver energy} \label{algo:nop_energy_cost}
\textbf{Input:} DNN structure, Chiplet count ($C$), Number of activations per layer ($A$), NoP bus width ($W$), Quantization bit ($Q_{bit}$), Energy per bit ($E_{bit}$) \\
\textbf{Output:} Energy for NoP driver ($E_{D}$)\\
\SetAlgoLined

\textbf{Initialize:} $E_{D} \leftarrow 0$

\For {$c = 1:|\mathcal{C}|$} {

Find index of source layer ($l$) \\
Find index of destination layer ($l+1$) \\

\tcc{\textbf{Number of packets between two consecutive chiplets}}
$N_{p} = \lceil \frac{A(l)Q}{W} \rceil $

$N_{bits} = N_p \times Q_{bit}$

$E_{D} = E_{D} + N_{bits} \times E_{bit}$

}

\end{algorithm}
%
Next, the engine evaluates the NoP transmitter/receiver (TX/RX) 
circuit\BT{s}, including the clocking circuitry.
\NC{The engine utilizes E$_{bit}$, number of TX/RX channels, bandwidth,} chiplet-to-chiplet data volume, and operation frequency to generate the energy and latency cost of the TX/RX 
circuit\BT{s}. 
Algorithm~\ref{algo:nop_energy_cost} details the energy calculation for the NoP driver.
We compute the total number of bits between chiplets. 
Further\BT{more}, we obtain the energy per bit ($E_{bit}$) from prior works, as shown in Figure~\ref{fig:crosssectional}(right).
We multiply the number of bits and energy per bit to obtain the total energy for TX/RX channel, as shown in line 9 of Algorithm~\ref{algo:nop_energy_cost}.
Next, the TX/RX circuit area from prior implementations (Figure~\ref{fig:crosssectional}) is utilized to \BT{obtain} 
the NoP driver area cost.
Finally, the NoP engine combines the performance metrics for the interconnect and the driver to generate the overall NoP performance.
\NC{We summarize the functional flow of the NoP engine:}
\begin{itemize}
    \item NoP trace generation based on the inter-chiplet layer partition, chiplet placement, and inter-chiplet data transfer volume.
    \item NoP interconnect evaluation using a cycle-accurate simulator to generate area, energy, and latency metrics.
    \item NoP TX/RX driver and router modeling based on real measurements. Finally, the NoP engine combines the interconnect and NoP driver metrics to generate the overall NoP performance.
\end{itemize}


\subsection{DRAM Engine}
\begin{figure}[t]
	\centering
	\vspace{-3mm}
	\includegraphics[width=1\textwidth]{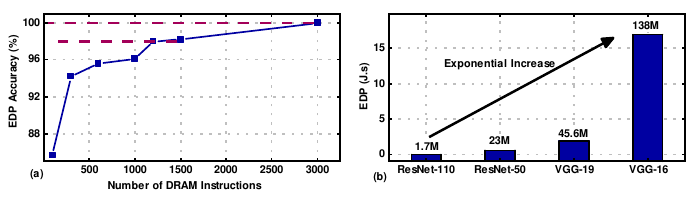}
	\vspace{-5mm}
	\caption{(a) \BT{The accuracy of EDP prediction}
	for different number\BT{s} of instructions processed to represent 3,000 DRAM instructions. Reduction in the number of instructions to half results in less than 2\% EDP accuracy degradation for half the simulation time, and (b) EDP of DRAM transactions \NC{(DDR4)} for different DNNs. There is an exponential increase in DRAM cost with an increase in DNN model size.}
	\label{fig:DRAM_req}
	\vspace{-5mm}
\end{figure}
The chiplet-based IMC architecture consists of a DRAM chiplet that acts as the external memory for the IMC chiplets.
The DRAM engine performs the external memory access estimation for the chiplet-based IMC architecture.
In this work, we assume that the DRAM only transfers the \BT{entire set of} weights to the chiplet one time before the inference task is performed.
\NC{Hence, it remains constant for a given DNN across different architectural configurations and inference runs.}

The engine consists of a DRAM request generator, RAMULATOR~\cite{kim2015ramulator} for estimating the latency for the DRAM transactions, and VAMPIRE~\cite{ghose2018your} to estimate the DRAM transaction power.
First, the choice of DRAM is determined based on the user input. 
Currently, SIAM supports both DDR3 and DDR4.
For DDR3 and DDR4, we incorporate the DRAM models detailed in~\cite{micron_ddr3_model, micron_ddr4_model}.
Next, for a given DNN model, the model size and data precision are determined from the user inputs.
Further\BT{more}, the DRAM engine generates the required traces and memory requests with time stamps. 
The requests include the location within the DRAM memory and the operation.

SIAM utilizes a customized version of the cycle-accurate simulator RAMULATOR~\cite{kim2015ramulator} and the model-based power analysis tool VAMPIRE~\cite{ghose2018your}. 
The customization includes the addition of support for larger DNNs, different data precision, and the addition of custom DDR3/DDR4 models.
Furthermore, to reduce \BT{the} simulation time for large DNNs such as VGG-16 (138M parameters), \NC{the DRAM engine breaks down the total number of instructions into smaller sets.}
\NC{The engine then performs the estimation for one set of instructions and multiplies it by the total number of sets required to represent all the weights in the DNN.}
To calibrate the method, we perform an experiment for 3,000 instructions broken down into a number of smaller sets of instructions.
Figure~\ref{fig:DRAM_req}(a) shows the corresponding energy-delay-product (EDP) accuracy for different sizes of instruction sets.
A reduction in 50\% of DRAM instructions to the engine results in less than 2\% EDP accuracy degradation than \BT{that} at 100\% instructions.
Furthermore, the reduced number of instructions allows for reduced simulation time for the DRAM engine.
We establish that, through this method, the DRAM engine performs fast and accurate estimation of external memory access for the entire range of DNNs.
Figure~\ref{fig:DRAM_req}(b) shows the overall EDP for different networks across different datasets for DDR4.
There is an exponential increase in EDP with the increase in the model size of the DNN.
\begin{figure}[t]
	\centering
	\includegraphics[width=0.9\textwidth]{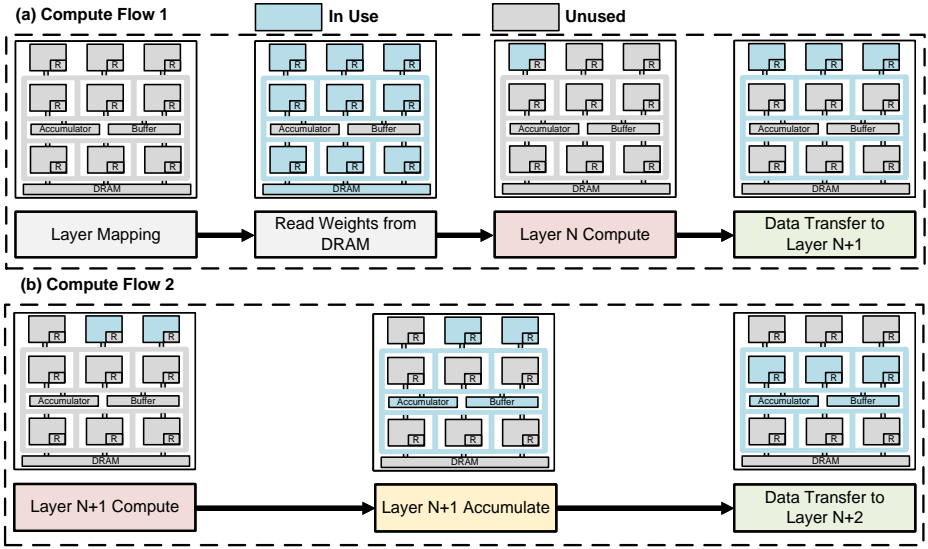}
	\vspace{-4mm}
	\caption{Computation dataflow within the chiplet-based IMC architecture in SIAM. Two cases arise, (a) no layer is partitioned across two or more chiplets, and (b) a layer is partitioned across two or more chiplets.}
	\label{fig:dataflow}
	\vspace{-4mm}
\end{figure}
To summarize, the following are the key steps in the execution of the DRAM engine:
\begin{itemize}
    \item Generate DRAM requests based on the DNN model size and data precision\BT{.}
    \item Calculate DRAM transaction latency cost using a customized version of RAMULATOR, 
    \BT{and calculate} the power consumption using a customized version of VAMPIRE\BT{.}
    \item Combine the outputs to generate the overall DRAM access cost.
\end{itemize}

\section{SIAM Dataflow}
\NC{This section presents the default dataflow \BT{in} 
the generated chiplet-based IMC architecture.}
Figure~\ref{fig:dataflow} shows \BT{an example of} the computation dataflow within the SIAM architecture.
Before performing the inference task, the weights are retrieved from the DRAM and mapped to the IMC chiplets based on the output from the partition and mapping engine (detailed in Section~\ref{sec:part_and_mapping}), as shown in Figure~\ref{fig:dataflow}(a).

Two cases can arise during the partitioning: first, no layer is distributed across two or more chiplets; second, a layer is distributed across two or more chiplets. 
The two cases result in two different scenarios within the execution dataflow.
Consider \BT{that} layer N of the DNN \BT{is} mapped onto the first chiplet in the architecture, as shown in Figure~\ref{fig:dataflow}(a).
During the computation, the entire layer is consumed within \BT{one} chiplet, 
producing the computed output activations from layer N.
Both the global accumulator and buffer are not utilized in the process and are turned off. 
After the computation, the output activations are transferred to the chiplets that implement layer N+1.
For layer N+1, \BT{let's assume that} two chiplets are \BT{required} 
to map the weights.
Hence, the NoP transfer\BT{s} the output activation from layer N to both 
chiplets housing layer N+1, as shown in Figure~\ref{fig:dataflow}(a).
Figure~\ref{fig:dataflow}(b) shows the computation flow for layer N+1.
Both 
chiplets perform the computation in a parallel manner. 
The mapping ensures that the same number of weights are mapped to each chiplet, thus avoiding the workload imbalance issue.
After completion of the computation, the generated partial sums are accumulated using the global accumulator and buffer. 
Then, the accumulated outputs from layer N+1 are transferred to the chiplets housing weights of layer N+2.
The process is repeated \BT{until} 
all the layers are completed and the final output is obtained. 
Algorithm~\ref{algo:data_flow} details the algorithmic implementation of the dataflow utilized \BT{in} 
the SIAM IMC chiplet architecture.
\begin{algorithm}[t]
\SetNoFillComment
\caption{Dataflow for SIAM IMC Chiplet Architecture} \label{algo:data_flow}
\textbf{Input:} Weights, input features, and total number of layers of the DNN \\
\textbf{Output:} Execution flow of the DNN on SIAM chiplet-based IMC architecture \\
\SetAlgoLined

 \While {i < Total number of layers}  {  
 \tcc{\textbf{Partitioning and mapping of the DNN}}
 Calculate number of chiplets required for $i^\mathrm{th}$ layer \\
 Perform computation for $i^\mathrm{th}$ layer \\
 
 \tcc{\textbf{Check if layer is distributed across chiplets}}
 \If {Number of chiplets > 1}{ 	
   Data transfer to accumulator \\
   Partial sum accumulation for $i^\mathrm{th}$ layer \\
}
Data transfer to chiplets of $(i+1)^\mathrm{th}$ layer \\
$i$ \textleftarrow $i+1$
}
\end{algorithm}
\section{Experimental Evaluation}\label{sec:expt}
\NC{We perform a wide range of experiments to 
\BT{demonstrate the effectiveness of} the proposed SIAM simulator.
These include detailed analysis of homogeneous and custom IMC chiplet architectures, comparison between monolithic and chiplet IMC architectures, calibration with real silicon data from SIMBA~\cite{shao2019simba}, comparison of the performance with GPUs, and evaluation of the SIAM's simulation time. 
We also illustrate the three characteristics of SIAM, flexibility, scalability, and simulation speed as shown below:
\begin{itemize}
    \item \textit{Flexibility and scalability:} Supporting different DNNs across datasets, two types of DNN partition to the IMC chiplets, and support for different IMC tile and chiplet configurations (Section~\ref{sec:custom_homo_imc}, Section~\ref{sec:monolithic_chiplet_comp}).
    \item \textit{Simulation speed:} Fast design space exploration of SIAM is demonstrated for different DNNs (Section~\ref{sec:simtime}).
\end{itemize}}
\subsection{Experimental Setup}
The DNNs that we evaluated include ResNet-110 (1.7M) on CIFAR-10, VGG-19 (45.6M) on CIFAR-100, ResNet-50 (23M) on ImageNet, and VGG-16 (138M) on ImageNet.
We use 8-bit quantization for the weights and activations and a 32nm CMOS technology node for the hardware.
We perform the experiments on an Intel Xeon CPU platform. 
The mapping of DNNs onto the IMC crossbars follows prior works~\cite{krishnan2020interconnect, shafiee2016isaac}.
Unless specified otherwise, all the experiments are performed based on the assumptions detailed next.
The chiplets are placed to achieve the least Manhattan distance. 
All results shown are for RRAM-based IMC architectures with the following parameters: one bit per RRAM cell, a R$_{off}$/R$_{on}$ ratio of 100, 16 tiles per chiplet, IMC crossbar size \BT{of} 128$\times$128, ADC resolution of 4-bits with 8 columns multiplexed, operating frequency of 1GHz~\cite{imani2019floatpim, shafiee2016isaac}, and a parallel read-out method.
We note that, the experiments do not consider the non-ideal effects within the RRAM-based IMC architecture.
The NoP parameters include a bandwidth of 250MHz, $E_{bit}$ of 0.54pJ/bit~\cite{poulton20130}, interconnect parameters such as width, thickness, and pitch from~\cite{poulton20130}, NoP TX/RX area of 5,304 $\mu$m$^2$~\cite{poulton20130}, NoP clocking circuit area of 10,609 $\mu$m$^2$~\cite{poulton20130}, and 32 channels (channel width).
\rev{We note that SIAM can support any NoP performance estimation as long as the NoP wiring parameters, bandwidth, channel width, TX/RX circuit area, clocking circuit area, and $E_{bit}$ are provided by the user.}
\NC{Finally, the reported results do not include the RRAM write, and DRAM read energy and latency.
Since we focus on DNN inference, the RRAM write and DRAM read operations are 
\YC{applied offline} before performing the inference. 
\YC{They do not involve in the inference runs and remain constant across}
IMC chiplet architectural configurations.}

\subsection{Custom and Homogeneous Chiplet-based IMC Design}\label{sec:custom_homo_imc}
\begin{figure}[t]
	\centering
	\includegraphics[width=0.5\textwidth]{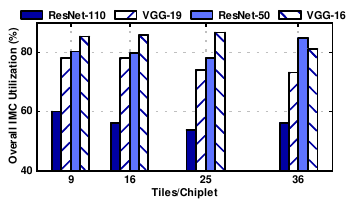}
	\vspace{-5mm}
	\caption{\rev{IMC utilization for a custom RRAM-based chiplet IMC architecture across different DNNs and different chiplet configurations. The mapping strategy adopted within SIAM ensures high utilization across all DNNs.}}
	\label{fig:utilization}
	\vspace{-5mm}
\end{figure}
\subsubsection{IMC Crossbar Utilization}

\rev{Figure~\ref{fig:utilization} shows the overall IMC crossbar utilization for a custom RRAM-based chiplet IMC architecture across different tiles per chiplet and DNNs.
We see that SIAM consistently achieves high (>50\%) IMC crossbar utilization.
The high IMC utilization indicates that mapping within SIAM generates chiplet-based IMC architectures that are area-efficient.
ResNet-110 has the lowest utilization due to the small network structure with fewer input and output features. 
At the same time, ResNet-50, VGG-19, and VGG-16 achieve $>$75\% utilization across the entire chiplet-based IMC architecture.
Hence, the partition and mapping engine within SIAM provides a flexible and efficient platform to generate chiplet-based IMC architectures with multiple configurations for design space exploration.}

\subsubsection{IMC Chiplet Performance Breakdown}
\begin{figure}[t]
	\centering
	\includegraphics[width=0.7\textwidth]{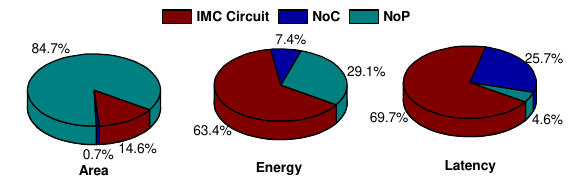}
	\vspace{-5mm}
	\caption{Breakdown of the different components contributing to the overall area, energy, and latency performance metrics, for a custom design RRAM-based chiplet IMC architecture when mapping ResNet-110 for CIFAR-10 dataset.}
	\label{fig:piechart}
	\vspace{-5mm}
\end{figure}
We analyze the breakdown of different components for the area, energy, and latency metrics for the RRAM-based chiplet IMC architecture. 
Figure~\ref{fig:piechart} shows the breakdown for \BT{the implementation of} ResNet-110 on CIFAR-10. 
We divide 
each metric into three main components, IMC circuit, NoC, and NoP. 
The IMC circuit component consists of the IMC crossbar array and associated peripherals, buffers (global and within chiplet), accumulators (global and within chiplet), pooling unit, and the activation unit.
\rev{At the same time, the NoP component consists of the NoP interconnect, NoP router, and the NoP driver and clocking circuit. }
Finally, the NoC component deals with the intra-chiplet interconnect and the NoC routers.

We first analyze the area metric. 
\rev{The NoP dominates the overall area with 84.7\%, while the NoC contributes the least to the area. The NoP drivers are designed such that differential signaling is utilized to avoid common-mode noise along with a clocking circuit for every N lanes. This results in increased circuitry for the TX-RX driver pairs and associated clocking circuitry for the 32 NoP channels. For example,~\cite{shao2019simba}  utilizes one clocking lane per 4 data lanes. The NoP router area depends on the technology node of the chiplet and the number of ports (default ports is set to 5). Simultaneously, the NoP link area depends on the wire properties. The NoP wire width and length are designed to maintain signal integrity at the specified frequency of operation. The wire for NoP link requires shielding on both sides of the signal, thus resulting in an increased pitch~\cite{poulton20130}. This results in a significant increase in the wiring area. We note that the NoP wire has a 56x larger metal pitch than that for the wires within the chiplet. Furthermore, an increased NoP channel width is needed for higher performance at the cost of increased area.}
For energy and latency, the IMC circuit component dominates with 63.4\% and 69.7\% contributions, respectively. 
The NoP contributes the second highest to energy, while the NoC contributes the least.
Simultaneously, NoC contributes the second highest to latency, while NoP contributes the least.
Overall, the area is \NC{dominated by the NoP, and the energy and latency are dominated by the IMC circuit.}

\subsubsection{\rev{NoP and NoC Performance Trade-offs}}
\begin{figure}[t]
	\centering
	\includegraphics[width=1\textwidth]{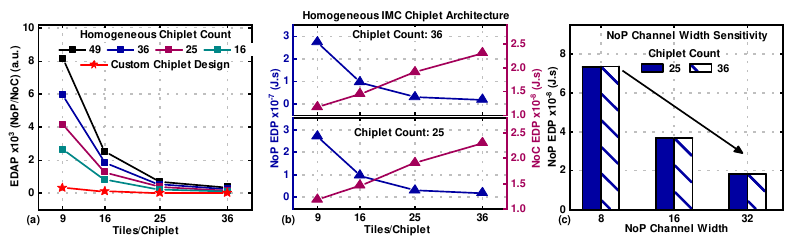}
	\vspace{-5mm}
	\caption{\rev{NoP and NoC trade-off analysis for ResNet-110 on CIFAR-10 dataset. (a) Ratio of the energy-delay-area product (EDAP) of NoP to NoC for both homogeneous and custom chiplet-based IMC architectures. The increase in tiles per chiplet reduces the NoP/NoC EDAP, (b) NoP and NoC energy-delay product (EDP) for a 36 chiplet count configuration of homogeneous RRAM-based chiplet IMC architectures. An increased tiles per chiplet leads to higher NoC cost and lower NoP cost.}}
	\label{fig:EDAP}
	\vspace{-4mm}
\end{figure}
\rev{We compare the EDAP for the NoC and NoP interconnect. 
Figure~\ref{fig:EDAP}(a) shows the ratio of the EDAP of the NoP to that of NoC for ResNet-110 on CIFAR-10, for both homogeneous and custom RRAM-based chiplet IMC architectures.
When there are fewer number of tiles per chiplet, more IMC chiplets are used to map the DNN to the IMC crossbars, resulting in distributed computing. 
This results in an increase in the data transfer volume across chiplets and higher NoP EDAP compared to that of NoC.
Furthermore, at higher chiplet counts, the NoP is much larger and results in increased area, thus increasing the EDAP.
As we increase the number of tiles per chiplet, computations are more localized, leading to reduced volume of data transfer across chiplets. 
This reduces the ratio of NoP EDAP to NoC EDAP.
The custom chiplet-based IMC architecture consists of the required number of chiplets to map the DNN under consideration.
In addition, the custom chiplet architecture is designed specific to a DNN, resulting in a highly localized computing platform with a smaller NoP.
Hence, the ratio of NoP EDAP to NoC EDAP is very small and is relatively insensitive to the change in tiles per chiplet.}

\rev{To further understand the trade-off between NoC and NoP, we evaluate the energy-delay product (EDP) of NoC and NoP separately.
Figure~\ref{fig:EDAP}(b) shows the energy-delay product (EDP) for the NoP and NoC for a 36 chiplet count configuration of homogeneous chiplet IMC architecture.
The x-axis shows the number of tiles in each chiplet. The EDP of NoP reduces with the increasing  number of tiles in each chiplet. The reduced EDP of NoP is achieved due to the highly localized computing resulting in lesser inter chiplet communication data volume.
At the same time, the NoC EDP increases with an increase in tiles per chiplet. The increased EDP is due to the larger NoC size (3x3 for 9 tiles per chiplet compared to 4x4 for 16 tiles per chiplet) and the increased intra-chiplet communication volume. Hence, a balance between the NoP and NoC cost is essential for optimal DNN inference performance with chiplet-based IMC architectures. We note that a similar trend is seen for other chiplet count configurations. From this experiment (ResNet-110 on CIFAR-10), we can conclude that the design with 16 tiles per chiplet provides a good balance in communication volume between NoC and NoP.
}

\subsubsection{\rev{Overall Hardware EDAP and Area}}
\begin{figure}[t]
	\centering
	\includegraphics[width=1\textwidth]{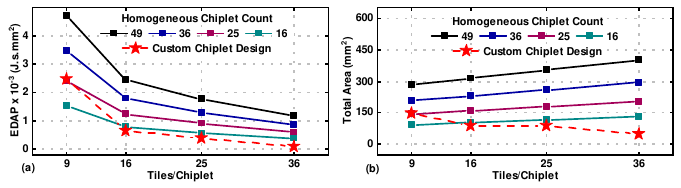}
	\vspace{-5mm}
	\caption{\rev{Energy-delay-area product as the metric. (a) Overall EDAP  and (b) total area for the homogeneous and custom RRAM-based chiplet IMC architecture when mapping ResNet-110 for CIFAR-10 dataset. The results indicate that a custom architecture outperforms a homogeneous architecture. The increased number of tiles per chiplet provides better performance at the cost of increased area for the homogeneous chiplet IMC architecture, while providing better performance and lower area for the custom chiplet IMC architecture.}}
	\label{fig:EDAP_new}
	\vspace{-5mm}
\end{figure}
\rev{Figure~\ref{fig:EDAP_new}(a) shows the overall performance (EDAP) of the RRAM-based chiplet IMC architecture for ResNet-110 on CIFAR-10.
For a homogeneous chiplet-based IMC architecture, the EDAP increases with higher chiplet counts (lower tiles per chiplet), resulting in higher chip area. 
Furthermore, with higher tiles per chiplet, the total energy contribution from the NoP reduces due to highly localized computing from the larger chiplet size and a lower NoP data volume. 
Overall, higher tiles per chiplet and lower chiplet count allow for a reduced EDAP in homogeneous RRAM-based chiplet IMC architectures.
Simultaneously, the custom chiplet architecture has better performance than the homogeneous architecture due to the reduced NoP size and a customized architecture for the given DNN.
A similar outcome arises for the custom design on increasing the tiles per chiplet, with the reduction in the EDAP.
}

\rev{Figure~\ref{fig:EDAP_new}(b) shows the overall area of the chiplet-based IMC architecture with different tiles per chiplet and different chiplet counts (homogeneous) and the custom chiplet architecture. The increase in the number of tiles per chiplet results in a higher area for the homogeneous chiplet architecture. The increase in area is due to the larger chiplet size while keeping the total chiplet count fixed. For example, the area for a 36 chiplet count and 16 tiles per chiplet architecture is larger than that of the 36 chiplet count and 9 tiles per chiplet architecture. Furthermore, the custom chiplet IMC architecture utilizes the required number of chiplet to map the whole DNN. Such an architecture benefits from increasing the tiles per chiplet since fewer chiplets are required to map the DNN. This results in lower NoP area and chiplet area (IMC circuit and NoC). Hence, with the increase in the number of tiles per chiplet the total area is reduced for the custom chiplet IMC architecture.
}

\subsection{Comparison between Monolithic and Chiplet-based IMC Architectures}\label{sec:monolithic_chiplet_comp}
\begin{figure}[t]
	\centering
	\includegraphics[width=0.9\textwidth]{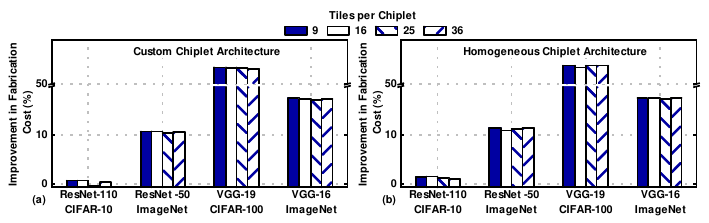}
	\vspace{-5mm}
	\caption{\rev{Improvement in fabrication cost (\$) for the RRAM-based chiplet IMC architecture, (a) custom and (b) homogeneous, as compared to a monolithic RRAM-based IMC architecture. Smaller DNNs like ResNet-110 have similar cost for both architectures, while larger DNNs such as VGG-19 have up to 60\% improvement.}}
	\label{fig:compare_single_chiplet}
	\vspace{-4mm}
\end{figure}
We perform a comparison between a custom monolithic RRAM-based architecture and both homogeneous and custom RRAM-based chiplet IMC architectures. 
Due to increased area, a monolithic IMC architecture suffers from increased defect ratio and lower yield. 
Consequently, a large monolithic IMC architecture experiences a very high fabrication cost, as shown in Figure~\ref{fig:first_fig}(a).
\rev{Figure~\ref{fig:compare_single_chiplet} shows the improvement in fabrication cost of (a) custom and (b) homogeneous RRAM-based chiplet IMC architectures, compared to that of a monolithic RRAM-based IMC architecture. 
We observe that the improvement is similar for different number of tiles per chiplet for a particular DNN. The increase in the number of tiles per chiplet results in a reduction in the total used chiplets, while keeping the same utilization across the used chiplets. Moreover, the improvement is similar for both custom and homogeneous chiplet architecture for a particular DNN. The improvement in fabrication cost is a strong function of the DNN structures. DNNs with fewer parameters exhibit less improvement. For example, ResNet-110 with 1.7M number of parameters shows up to 0.57\% improvement in fabrication cost. At the same time, VGG-19 with 45.6M number of parameters show more than 50\% improvement in the fabrication cost. The reduced fabrication cost is attributed to lower defect ratio and increased yield achieved through smaller chiplets connected together to form a large system. Hence, larger and branched DNNs benefit significantly from chiplet-based IMC architectures.}

\subsection{Calibration with SIMBA}
\begin{figure}[t]
	\centering
	\includegraphics[width=1\textwidth]{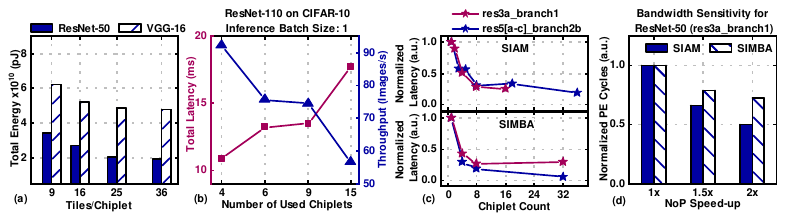}
	\vspace{-5mm}
	\caption{(a) Total energy for DNN inference reduces with an increase in \BT{the} number of tiles per chiplet (chiplet size), (b) Total inference latency and throughput for ResNet-110 on CIFAR-10. Due to the small network size, a lower number of chiplets provide better performance, \NC{(c) Normalized latency for two layers within ResNet-50 on ImageNet for SIAM RRAM-based chiplet IMC architecture and SIMBA~\cite{shao2019simba}. The decreasing trend of latency with increasing chiplet count exhibited by SIAM is consistent with SIMBA, (d) Bandwidth sensitivity for a layer within ResNet-50. The decreasing trend in the PE cycles with increasing NoP speed-up similar to SIMBA.}}
	\label{fig:SIBA_Calib}
	\vspace{-4mm}
\end{figure}
\NC{This section presents the calibration results for the SIAM RRAM-based chiplet IMC architecture compared to published silicon data from SIMBA~\cite{shao2019simba}.
We note that, there is no prior work that reports real silicon data for chiplet-based IMC architectures.
Therefore, we choose SIMBA, the work that has the most resemblance to that of SIAM. 
We utilize the NoP driver circuit and the signaling technique similar to \BT{those} 
in SIMBA for our experiments.
Furthermore, the NoP interconnect parameters utilized in SIAM are closed to that in SIMBA.}

\NC{\noindent\textbf{Total Energy: }Figure~\ref{fig:SIBA_Calib}(a) shows the total energy for inference across different number of tiles per chiplet for both ResNet-50 and VGG-16 on the ImageNet dataset.
The increase in \YC{the} number of tiles per chiplet results in a reduction in the total \YC{number of} chiplets used to map the DNN and, in turn, a reduction in the total inference energy. 
The same trend is reported in SIMBA (ResNet-50).
}

\NC{\noindent\textbf{Total Latency: }We evaluate the effect of chiplet scaling or, in other words, the number of chiplets used to map a small DNN. Figure~\ref{fig:SIBA_Calib}(b) shows the total inference latency and throughput for ResNet-110 on CIFAR-10 dataset.
Since ResNet-110 is a small DNN, distributing the computation across more chiplets results in a sub-optimal configuration.
A similar trend is shown in SIMBA for a small DNN, DriveNet~\cite{bojarski2017explaining}.
}

\NC{\noindent\textbf{Layer Sensitivity:} We consider two representative layers in ResNet-50, res3a\_branch1 and res5[a-c]\_branch2b , same as that shown in SIMBA.
We analyze the latency 
\skm{(normalized with the latency of the design consisting of only one chiplet)}
by varying the number of chiplets used to map the DNN layer.
We note that the chiplet count to map the DNN is different from SIMBA due to the difference in the computation element in SIAM (IMC crossbars) and SIMBA (MAC arrays).
The analysis shown in Figure~\ref{fig:SIBA_Calib}(c) (top) reveals that there is a decreasing trend in latency with increasing number of chiplets.
For res3a\_branch1 the latency reduces initially with the increase in chiplet count and finally increases slightly \skm{(with chiplet count of 16)}. 
The increase in latency is due to a higher NoP latency from 
\YC{more} distributed 
computation. 
res5[a-c]\_branch2b shows a consistent decrease in the latency with the increase in chiplet count.
The\YC{se} trend\YC{s} 
\YC{are} consistent with that reported in SIMBA, as shown in Figure~\ref{fig:SIBA_Calib}(c) (bottom).}

\NC{\noindent\textbf{PE cycles vs NoP speed-up:} In this experiment, we vary NoP frequency and analyze the variation in PE cycles of res3a\_branch1 layer in ResNet-50. 
We normalize it to the 1$\times$ case 
\skm{to be consistent} with SIMBA.
The SIAM chiplet IMC architecture shows decreasing PE latency with increasing NoP bandwidth, which conforms SIMBA, as shown in Figure~\ref{fig:SIBA_Calib}(d).}

\NC{
\YC{In summary, these comparisons confirm that, during the scaling of chiplet parameters, such as the number of chiplets and their utilization, SIAM predicts similar trends as} 
the measured results from real silicon.}

\subsection{Comparison with GPUs}
We compare the performance of the chiplet-based IMC architecture generated using SIAM with state-of-the-art GPUs such as Nvidia V100 and T4.
Unlike GPUs, SIAM generates architectures \BT{for} energy-efficient inference \BT{using small batch sizes}.
All GPU hardware performance numbers have been adopted from those reported in~\cite{shao2019simba}.
We compare the performance of the architecture for inference with a batch size of one.
For ResNet-50 on ImageNet dataset, the architecture generated using SIAM (36 tiles per chiplet) results in a total area of 273 mm$^2$ as compared to 525 mm$^2$ for T4 and 815 mm$^2$ for V100.
The reduced area is attributed to the high compute density achieved using an IMC design and the support for a wide range of computations within the GPU.
We also compare the \BT{energy-efficiency of} SIAM IMC architecture to that of the GPUs for ResNet-50 \NC{on ImageNet.} 
SIAM achieves 130$\times$ and 72$\times$ higher energy-efficiency as compared to V100 and T4 GPUs, respectively.
\rev{The chiplet-based IMC architecture has higher performance since IMC architectures have all weights on chip, thus avoiding the external memory access. Furthermore, IMC utilizes analog domain computation within the crossbar arrays that are more energy-efficient than regular multiply-and-accumulate (MAC) units~\cite{shafiee2016isaac}.} 


\subsection{Simulation Time}\label{sec:simtime}
%

\begin{table}[t]
\centering
\caption{Simulation Time for SIAM}
\label{tab:sim_time}
\resizebox{0.6\textwidth}{!}{%
\begin{tabular}{@{}c|c|c|c@{}}
\toprule
\multirow{2}{*}{\textbf{Network}} & \multirow{2}{*}{\textbf{Dataset}} & \multirow{2}{*}{\textbf{Model Size (M)}} & \multirow{2}{*}{\textbf{Simulation Time (Hours)}} \\
           &           &      &      \\ \midrule
ResNet-110 & CIFAR-10  & 1.7 & 0.2  \\ \midrule
VGG-19     & CIFAR-100 & 45.6 & 0.36 \\ \midrule
ResNet-50  & ImageNet  & 23   & 1.26 \\ \midrule
VGG-16     & ImageNet  & 138  & 4.26 \\ \bottomrule
\end{tabular}%
}
\end{table}
Table~\ref{tab:sim_time} shows the simulation time for the proposed chiplet-based IMC simulator, SIAM, for different DNNs across different datasets.
The simulation time is extracted by running SIAM on an Intel Xeon W-2133 CPU platform with 12 cores and 32GB RAM.
The range of the simulation times varies from a couple of minutes for small DNNs to a few hours for large DNNs.
For a fair analysis, we report the overall simulation time that includes the partitioning and mapping, circuit and NoC simulation, NoP estimation, and DRAM access estimation.
For example, ResNet-110 with 1.7M parameters for CIFAR-10 dataset takes 12 minutes (0.2 hours) for SIAM to perform the performance benchmarking.
A large DNN such as VGG-16 with 138M parameters on the ImageNet dataset takes 4.26 hours for SIAM to perform the benchmarking.

\rev{Finally, we perform a comparison between SIAM and NeuroSim~\cite{peng2019dnn+} in terms of simulation for benchmarking a RRAM-based monolithic IMC architecture. 
We note that we choose a monolithic IMC architecture as no other simulator supports chiplet-based IMC architecture benchmarking.
We perform the comparison for four networks namely, ResNet-110 (1.7M), VGG-19 (45.6M), ResNet-50 (23M), and VGG-16 (138M).
For ResNet-110 SIAM requires 60s while NeuorSim takes 30s while for VGG-19, SIAM takes 86s and NeuroSim takes 46s. 
Furthermore, for ResNet-50 SIAM takes 818s while NeuroSim takes 276s and for VGG-16 SIAM takes 3110s while NeuroSim takes 1110s. 
We note that the simulation times for SIAM are in the same range as that of NeuroSim rather than being orders of magnitude higher. 
The increased simulation times are due to the additional functionality that SIAM provides in NoC and DRAM performance estimation.}
\section{Conclusion and Discussion}
This work presents SIAM, a novel performance benchmarking tool for chiplet-based IMC architectures.
To the best of our knowledge, this will be the first
benchmarking tool for design space exploration \BT{of chiplet-based IMC architectures}.
SIAM integrates device, circuits, architecture, NoC, NoP, and DRAM estimation into an end-to-end system. 
It supports two types of chiplet architectures, homogeneous and custom, generated using different partition schemes.
In addition, SIAM supports different DNNs across different datasets and \BT{various} 
IMC chiplet configuration\BT{s}.
Through this \BT{study}, we establish the scalability and flexibility features of SIAM.
We demonstrate the speed of SIAM by evaluating the simulation time of SIAM for different DNNs and comparing it against state-of-the-art chiplet simulators like NeuroSim for monolithic IMC architectures.
Next, we calibrate SIAM with respect to publish\BT{ed} silicon result, SIMBA, 
\BT{which confirms} that \BT{the trends projected by SIAM} 
match the measured results from real silicon.
Finally, we compare the performance of the generated chiplet-based IMC architecture using SIAM to that of state-of-the-art GPUs.
The SIAM chiplet architecture achieve\BT{s} 130$\times$ and 72$\times$ improvement in energy-efficiency compared to Nvidia V100 and T4 GPUs, respectively.
\section*{Appendix A}\label{sec:appendix}

\noindent\textbf{Estimating manufacturing cost of a chip:}
%
Let us consider a reference chip with area $A_{ref}$ and $N_{ref}$ chips per wafer.
The cost of the reference chip ($C_{ref}$) is expressed as shown in Equation~\ref{eq:chip cost}.
\begin{equation} \label{eq:chip cost}
    \small C_{ref} = \frac{C_{Total}}{\eta_{ref} N_{ref}}
\end{equation}
where, $\eta_{ref}$ is the yield of the wafer.
Number of chips per wafer ($N_{ref}$) is expressed in~\ref{eq:num_chips}~\cite{chip_cost}.
\begin{equation} \label{eq:num_chips}
    \small N_{ref} = D\pi \Big( \frac{D}{4A_{ref}} - \frac{1}{\sqrt{2A_{ref}}} \Big)
\end{equation}
where $D$ is the wafer diameter.
\NC{The cost of a target ($C_{target}$) and normalized cost ($C_{norm}$) are:
\begin{equation}
    \small C_{target} = \frac{C_{Total}}{\eta_{target} N_{target}} \small C_{norm} = \frac{C_{target}}{C_{ref}} = \frac{N_{ref} \eta_{ref}}{N_{target} \eta_{target}}
\end{equation}
}
%
Assuming poisson defect model, $\eta = e^{-D_0 A}$, where $D_0$ is the defect density.
Replacing the expression of $\eta$ in \NC{expression for $C_{norm}$}, we obtain:
\begin{equation}
    \small C_{norm} = \frac{N_{ref} e^{-D_0 A_{ref}}}{N_{target} e^{-D_0 A_{target}}} = \frac{N_{ref}}{N_{target}} e^{-D_0 (A_{ref} - A_{target})}
\end{equation}

\noindent\textbf{Verification of chip cost estimation:}
To verify the chip cost estimation, we assume $A_{ref} = 296mm^\mathrm{2}$, $D_0 = 0.012/mm\mathrm{^2}$ and $D=152.4mm$.
The comparison reveals that the estimation is 98\% accurate \NC{with respect to the real chip cost of commercial processors~\cite{tam2018skylake}}.

\camready{\section{Acknowledgements}
This work was supported by C-BRIC, one of six centers in JUMP, a Semiconductor Research Corporation (SRC) program sponsored by DARPA, 
and SRC GRC Task 3012.001.}
\bibliographystyle{ACM-Reference-Format}
\bibliography{2.5D_DNN_sim}
\end{document}